\documentclass[10pt,twocolumn,letterpaper]{article}

\usepackage{iccv}
\usepackage{times}
\usepackage{epsfig}
\usepackage{graphicx}
\usepackage{amsmath}
\usepackage{amssymb}

\newcommand{\txt}[1]{{\texttt{#1}}}

\newcommand{\T}[1]{{\mathcal{#1}}} 
\newcommand{\V}[1]{{\mathbf{#1}}} 
\usepackage{graphicx}
\usepackage{enumitem}
\usepackage{comment}
\usepackage{enumitem}
\usepackage{wrapfig}
\usepackage{xcolor}
\usepackage{booktabs}
\usepackage{subfig}

\usepackage{subfloat}

\newcommand{\ours}{\textsc{KIDD}}
\newcommand{\ourstit}{Koopman-based Interpretable Decomposition and Disentanglement}

\usepackage[pagebackref=true,breaklinks=true,letterpaper=true,colorlinks,bookmarks=false]{hyperref}

\iccvfinalcopy 

\def\iccvPaperID{10867} 

\begin{document}

\title{Self-Supervised Decomposition, Disentanglement and Prediction\\ of Video Sequences while Interpreting Dynamics: \\A Koopman Perspective}


\author{Armand Comas 
\and
Sandesh Ghimire
\and
Haolin Li 
\and
Mario Sznaier
\and
Octavia Camps 
\and
\small \texttt {\{comasmassague.a, s.ghimire, li.haolin, m.sznaier, o.camps\} @northeastern.edu}\\
Computer Engineering,
Northeastern University, Boston\\
}

\maketitle


\begin{abstract}
Human interpretation of the world encompasses the use of symbols to categorize sensory inputs and compose them in a hierarchical manner. One of the long-term objectives of Computer Vision and Artificial Intelligence is to endow machines with the capacity of structuring and interpreting the world as we do. Towards this goal, recent methods have successfully been able to decompose and disentangle video sequences into their composing objects and dynamics, in a self-supervised fashion. However, there has been a scarce effort in giving interpretation to the dynamics of the scene. We propose a method to decompose a video into moving objects and their attributes, and model each object's dynamics with linear system identification tools, by means of a Koopman embedding. This allows interpretation, manipulation and extrapolation of the dynamics of the different objects by employing the Koopman operator $\mathcal{K}$. We test our method in various synthetic datasets and successfully forecast challenging trajectories while interpreting them.
\end{abstract}

\section{Introduction}
Unsupervised learning of symbolic representations from high dimensional data poses a great challenge to current machine intelligence. As humans, our intuitive modeling of the world is based on abstract categories, or symbols. 
The universe of those categories is unbounded and continuous. Fortunately, we can approach symbolic reasoning by discretizing and simplifying those categories. 
One interesting direction for categorization is based on compositionality, specialization and hierarchy.
Different concepts will be in charge of different tasks, and the hierarchical composition of their outputs will generate the complex behaviors we wish to model.

More specifically, we look at the task of visual perception. For visual scenes, a simplification of the symbols encompasses entities, their attributes and their interactions with the environment. Numerous efforts have focused on decomposing a static scene into its composing objects and background in an unsupervised fashion \cite{air, iodine,slot-attention, Burgess2019MONetUS, Lin2020SPACEUO}. In this case the objective is to learn the category of ``object" and its related attributes such as ``location", ``appearance" or ``depth", without labeling any of those categories. 
The supervision signal is often a surrogate task of coherence, such as hierarchical rendered reconstruction of the scene or imposing rules on the intermediate representations (\text{i.e.} contrastive learning \cite{cswm}).

Similarly, adding one dimension to the problem, many approaches have tackled unsupervised video decomposition \cite{sqair, drnet, dive, ddpae, stove, He2019TrackingBA, scalor}. 
An added issue in this case is finding the correspondences of the decomposed objects across time (\text{i.e.} tracking) while defining what an object is.
And as there is time, there is a future. Therefore, the question of how the scene \textit{will} look arises. Recurrent Neural Networks (RNNs) provide a useful class of models for forecasting tasks. Once a scene is decomposed, they are often used to model the underlying dynamics. However, RNNs suffer from exploding and vanishing gradient and it is hard to incorporate high-level constraints to the model. Related to the latter, an other issue of such methods is their lack of high-level interpretability. 

In this work, we advocate that a data-driven physics-based approach can bring a principled and interpretable perspective to the dynamics modelling, while preserving the model's predictive power. 
We make use of Koopman theory, which is based on the insight that a finite-dimensional nonlinear system can be transformed to an infinite-dimensional linear dynamical system, and then propagated in time using a linear operator $\mathcal{K}$. Therefore, we can apply tools of linear algebra and spectral theory, and the dynamical system can be understood as a composition of first-order impulse responses. Koopman theory has been successfully applied to model time series with many applications \cite{Schmid2008DynamicMD, Williams2015ADA, Lusch2018DeepLF}. We develop this further in Section \ref{sec:background}.

Our model, which we call  \ourstit{} (\ours{}),  (\textbf{i}) uses an attention-based tracking method to learn representations from video factorized into moving objects and their attributes: appearance, confidence and pose; (\textbf{ii}) finds a non-linear mapping to a the Koopman space for the dynamic latent representations; (\textbf{iii}) learns the Koopman operators that characterize the underlying dynamics of the training data; and (\textbf{iv}) performs unsupervised video prediction using the latent representations. In our experiments, we also propose simple decomposition techniques to interpret the objects dynamics.

\begin{figure*}[t]
    \centering
    \includegraphics[trim=150 30 150 575,clip,width=\textwidth]{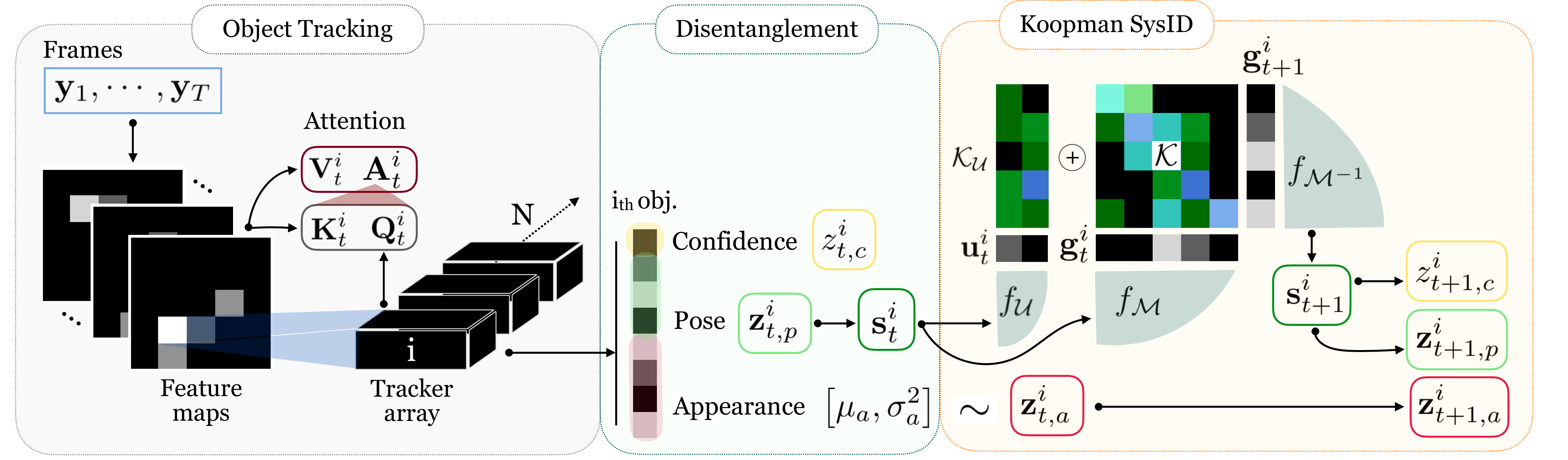}
    \caption{Overall architecture of \ours{}. In the left, an attention-based recurrent tracker decomposes the scene into its composing objects. In the center, the object representations are disentangled into Confidence, Appearance and Pose. In the right, the dynamic features of Pose are modelled and forecasted by using a Koopman embeding. These latent representations are later used to reconstruct or predict frames.
    }\label{fig:arch_all}
\end{figure*}

\section{Background}\label{sec:background}
We consider a time-invariant autonomous dynamical system of a single object on $\Re^n$ of the form:
\begin{equation}
    \V{s}_{t+1} = \Phi \left( \V{s}_{t}\right) \label{eq:nonlinearsystem}
\end{equation}
where $\V{s}_{t} \in \Re^n$ is the state of the system at time $t$. $\Phi: \Re^n \rightarrow \Re^n$ is a potentially non-linear function that defines the temporal transition of the states.

The fundamental insight of Koopman operator theory is that the finite-dimensional nonlinear dynamics of Equation \ref{eq:nonlinearsystem} can be transformed to an infinite-dimensional linear dynamical system by considering an appropriately chosen Hilbert space of scalar observables $g_j= \psi_j\left( \V{s}_{t}\right) \in \Re$ \cite{Koopman1931HamiltonianSA, Mezic2005SpectralPO}. 
The eigenfunctions $\psi_j\left( \V{s}_{t}\right):  \Re^n \rightarrow \Re$ of the Koopman operator are difficult to find, and some algorithms have been proposed to tackle the challenge. The most widely used are the dynamic mode decomposition (DMD) \cite{Schmid2008DynamicMD} and its extension to nonlinear observables, the extended DMD (EDMD) algorithm \cite{Williams2015ADA}.

Previous research used hand-crafted eigenfunctions to model the observable space. Those were chosen from function families or directly from previous knowledge of the physics of the problem. Currently, some approaches use deep neural networks to represent the observable space \cite{Lusch2018DeepLF, Morton2019DeepVK, Li2020Learning, Xiao2021CKNetAC, pmlr-v119-azencot20a}. Neural networks have the advantage of being universal approximators, and are effective in finding the Koopman invariant subspace given a downstream task.
With this latter perspective, Koopman methods have been applied successfully to fluid dynamics \cite{NEURIPS2018_2b0aa0d9, pmlr-v119-azencot20a}, atomic and molecular scale dynamics \cite{Xie2019GraphDN, Mardt2017VAMPnetsFD}, chaotic systems \cite{Brunton2017ChaosAA} or traffic dynamics \cite{Xie2019GraphDN}, between others.

Koopman methodology is data-driven, model-free and can discover the underlying dynamics and control of a given system from data alone \cite{Proctor2018GeneralizingKT}. The system identification problem is then reduced to finding the operator $\mathcal{K}$. This is usually done by linear regression given historical data (e.g. by means of ordinary least squares) or by end-to-end gradient-descent-based optimization. For the latter, the Koopman operator is often learned jointly with the mapping and inverse mapping.

The Koopman operator $\mathcal{K}: \Re^m \rightarrow \Re^m$ is defined by: 
\begin{equation}
        f_{\mathcal{\psi}} \circ \Phi \left( \V{s}_{t}\right)=\mathcal{K}f_{\mathcal{\psi}}\left( \V{s}_{t}\right),
\end{equation}
where $f_{\psi}: \Re^n \rightarrow \Re^m$ is the mapping from the state space to the observable space $\V{g}_t\in \Re^m$ and $\circ$ denotes the composition operator. Therefore:
\begin{align}
    &\V{g}_{t} = f_{\mathcal{\psi}}\left( \V{s}_{t}\right) \label{eq:mapping}\\
    &\V{g}_{t+1} = \mathcal{K}\V{g}_{t} \label{eq:koop-op}\\
    & \mathcal{K}\psi_j\left( \V{s}_{t}\right) = \lambda_j\psi_j\left( \V{s}_{t}\right),
\end{align}
where $\lambda_j \in \mathbb{C}$ is the eigenvalue of $\mathcal{K}$ corresponding to the eigenfunction $\psi_j$.
In some cases, Koopman is employed in presence of control inputs. There are different approaches to introducing inputs to Koopman (\textit{e.g.} \cite{Proctor2018GeneralizingKT, Li2020Learning}). In the studied cases, inputs model forces external to an object, originated from object-environment interactions. Therefore, inputs will depend both on the state of the object and the environment's geometry. The latter is learned implicitly from data. Consequently, Equations \ref{eq:mapping} and \ref{eq:koop-op} in the presence of inputs are modified as follows:
\begin{align}
    &\V{g}_{t} = f_{\mathcal{\psi}}\left(\V{s}_{t}\right), \V{u}_{t} =  f_{\mathcal{U}}\left(\V{s}_{t}\right) \label{eq:mapping-inp}\\
    &\V{g}_{t+1} = \mathcal{K}\V{g}_{t} +  \mathcal{K_\mathcal{U}}\V{u}_{t} \label{eq:koop-op-inp}
\end{align}
Here, $\V{u}_t$ depends on the current state $\V{s}_{t}$ (Closed-loop control). It is expected to be sparse, and low-dimensional. $\mathcal{K_\mathcal{U}}: \Re^u \rightarrow \Re^m$ is the input Koopman operator. We usually define dimension $u$ such that $u << m$.
A challenge will be to discover $\V{u}_t$ and the correct mapping to the Koopman manifold simultaneously.


\section{Related Work}
\textbf{Object decomposition}
Numerous recent publications have focussed on unsupervised decomposition of a scene into the different objects that compose it. Attend, Infer, Repeat (\textsc{AIR})\cite{air} presents an unsupervised way to count, locate objects and reconstruct a scene by providing structure in the inference module. AIR also infers the number of objects present in the scene. 
Following the path of structured design, \textsc{iodine} \cite{iodine} also presents an unsupervised way to decompose scene into multiple objects using  Gaussian mixture model as the generative model and amortized iterative inference. Leveraging  compositional structure it learns disentangled, interpretable and generalizable representations.
Similarly, slot attention \cite{slot-attention} introduces a general purpose plug-in network based on attention that discovers objects in an image. They draw lines with transformers and soft K-means clustering. One interesting feature of this model is that it can generalize to previously unsen composition and more objects since slots are not associated with objects, rather each slot has ability to capture any of the object.

\textbf{Video decomposition}
\textsc{drnet}\cite{drnet} is an early work that decomposes video into a static component (content) and a dynamic component (pose). This is a key idea that will be embraced by many of the discussed works. In order to achieve this decomposition, it makes use of an adversarial loss to enforce that the dynamic component doesn't carry identity information. \textsc{sqair} \cite{sqair} extends the idea of \textsc{air} to infer a video sequence by considering its temporal progression. Since different objects are discovered sequentially, \textsc{sqair} is not scalable. \textsc{scalor} \cite{scalor} overcomes this limitation by massive parallelization and is therefore able to handle hundreds of objects in a scene and predict their future trajectories simultaneously. Similar to \textsc{drnet}, \textsc{ddpae}\cite{ddpae}, decomposes a video into pose and content, where the content is static and the pose is dynamic and modeled using RNNs. \textsc{stove} \cite{stove} again decomposes and disentangles the objects in the latent space. It uses a graph to model the interaction between them. By using a Markov model in the latent space, \cite{stove} applies inference in the series. They show that their model generates realistic frames and conserves kinetic energy even when predicted for long time. Similarly, C-SWM \cite{cswm} uses graph neural network to learn relations between objects in a self-supervised way directly from raw videos by using structured model. However, in this case the supervision is done by means of contrastive learning and a bipartite loss. Tracking by Animation \cite{He2019TrackingBA} shows that decomposition, disentanglement and deterministic generation of objects is a good self-supervision signal for learning tracking. We will use their ideas on attention-based tracking for our method. 
\\
\\
\\

\textbf{Koopman Operator} Koopman Operator Theory studies transformation of nonlinear dynamics into a space with linear dynamics given by linear operator $\mathcal{K}$. It has been successfully used to disentangle the dynamic modes in complex dynamical systems using dynamic mode decomposition techniques like DMD \cite{Schmid2008DynamicMD}, and Extended DMD (EDMD)\cite{Williams2015ADA}. 
By leveraging the fact that Koopman operator based methods are completely data-driven and require a rich family of function to generate a mapping, many modern works have used deep neural networks to approximate the eigenvectors associated to the Koopman operator. This has resulted into several deep network based Koopman methods \cite{Lusch2018DeepLF, pmlr-v119-azencot20a, Morton2019DeepVK, Li2020Learning, Xiao2021CKNetAC, lran}.
This idea has recently found successful application in fluid dynamics \cite{NEURIPS2018_2b0aa0d9, pmlr-v119-azencot20a}, atomic and molecular scale dynamics \cite{Xie2019GraphDN, Mardt2017VAMPnetsFD}, chaotic systems \cite{Brunton2017ChaosAA} and traffic dynamics \cite{Xie2019GraphDN}.
The original Koopman operator theory is developed without any external inputs to the dynamical system. Later, it has been generalized to considering inputs \cite{Proctor2018GeneralizingKT}. This gives rise to other methods like Compositional Koopman \cite{Li2020Learning}, that have used Koopman theory to model object dynamics and interactions with other objects and the environment in a compositional way. The introduction of inputs to the dynamics modelling allows for applications in control and reinforcement learning. Our work uses ideas from the recent developments in Koopman-based modelling and self-supervised video decomposition and prediction to propose a joint decomposition of the static and dynamic components of a scene. This will allow for compositional generation and interpretability of the dynamics. To the best of our knowledge, this is the first work that tackles this set of problems jointly with an end-to-end model.

\section{Method}
Video often presents multiple objects in motion that generate a complex dynamical scene. In the pixel space, dynamics are strongly non-linear. But by decomposing the scene into abstract categories, dynamics are simpler to model and more interpretable. For our approach, we decompose the scene into its moving objects. For simplicity, we assume that there is no background.
We track and identify $N$ objects across input frames, and assign $3$ sets of variables to each one of them. Each object representation will be disentangled into the following categories:
\begin{itemize}
    \item \textbf{Pose} $\V{z}_{t,p} \in \Re^{4}$: Indicates the parameters for an affine spatial transformation of an object; $x$ and $y$ coordinates of the centroid, scale and ratio.
    \item \textbf{Appearance} $\V{z}_{t,a} \in \Re^{A}$: Modeled either dynamic across frames or static, a vector containing information about an objects appearance. 
    \item \textbf{Confidence} $z_{t,c} \in [0,1)$: A probability scalar indicating the certainty of an object being correctly modeled.
\end{itemize}
Our method uses concepts of soft-attention for feature-based tracking. We build on top of \cite{He2019TrackingBA} for our tracking mechanism, and modify the architecture to allow stochasticity and forecasting. Following, we describe the main modules that form our model:
\vspace{-2mm}
\paragraph{Tracking}
We encode each frame of a video $(\V{y}_1,\cdots, \V{y}_T)$ with a convolutional encoder, and obtain a feature map as $\V{H}_{t,y} = f_{\txt{enc}}\left( \left[\V{y}_t, \textsc{pe}\right] \right)$, where $\textsc{pe}$ is a positional encoding. We then track objects across features by using an array of trackers. $N$ trackers are initialized, where $N$ is an upper-bound on the expected number of objects in every scene. We define the tracker recurrent updates as:
\begin{align}
    &\V{h}^i_{t,tr} = f_{\txt{tr}}\left( \V{h}^i_{t-1,tr}, \V{H}_{t,y}\right)\label{eq:tracker_state}\\
    &\left[z^i_{t,c},\V{z}^i_{t,p}, \V{o}^i_{t,a} \right] = \texttt{FC}\left( \V{h}^i_{t,tr}\right)\\
    &\V{z}^i_{t,a} \sim \T{N}(\V{\mu}_{a}, \V{\sigma}_{a}^{2}), \quad
\left[\V{\mu}_{a}, \V{\sigma}_{a}^{2}\right] = \texttt{FC}(\V{o}^i_{t,a}), \nonumber
\end{align}
where  $f_{\txt{tr}}\left(\cdot\right)$ is an attention-based tracking function described in Equations \ref{eq:attn} and \ref{eq:tr-update}. We implement the appearance latent vector $\V{z}^i_{t,a}$ to be the only stochastic variable in this setup, given its inherent complexity. It is sampled from a gaussian distribution and trained as in a VAE framework.

\vspace{-2mm}
\paragraph{Tracker updates}
We use a content-based soft-attention mechanism \cite{Bahdanau2015NeuralMT}, relying on the Query, Key, Value triad. We attend to the information of the feature map $\V{H}_{t,y}$ that describes the $i_{\text{th}}$ object:\\
\begin{align}
    &\left[\V{Q}^i_{t}, \hat{\beta}^i_{t} \right] = \txt{FC}\left(\V{h}^i_{t-1,tr}\right),\text{ }  \beta^i_{t} = 1 + \text{ln}\left(1+e^{\hat{\beta}^i_{t}}\right),\nonumber\\& \V{K}^i_{t} = \V{H}^i_{t,y}, \quad \V{V}^i_{t} = \V{H}^i_{t,y} \nonumber \\ 
    &\V{u}^i_{t} =\V{A}^i_{t}\V{V}_{t}, \quad \V{A}^i_{t} = \txt{Softmax}\left( \beta^i_{t}\V{Q}^i_{t}\V{K}^T_{t}\right) \label{eq:attn}\\
    &\V{h}^i_{t,tr} = \txt{GRU}_{\txt{tr}}\left( \V{h}^i_{t-1,tr}, \V{u}^i_{t}\right)\label{eq:tr-update}
\end{align}
A \txt{GRU} \cite{Cho2014LearningPR} cell is used to update the hidden state of each tracker $\V{h}^i_{t,tr}$. Such state queries the features $\V{H}_{t,y}$. Note that the Value $\V{V}^i_{t}$ and Key $\V{K}^i_{t}$ are both the raw convolutional features, modified iteratively following Equations \ref{eq:ew} and \ref{eq:write}. $\beta^i_{t} \in (1, \inf)$ is the Query strength. An attention map $\V{A}_t^i$ is applied to the values resulting in the current input $\V{u_t^i}$ to the \txt{GRU}.

\paragraph{Memory}
We need a mechanism to provide information across trackers, while preserving the identity of each object through time. Based on \cite{He2019TrackingBA, Graves2016HybridCU}, trackers interact through external memory by using interface variables. We use the frame convolutional features $\V{H}_{t,y}$ as external memory, and implement read and write operations. When iterating through trackers, the input $\V{H}_{t,y}$ will be updated as follows:
\begin{align}
    &\V{e}^i_{t} = \txt{Sigmoid}\left(\hat{\V{e}}^i_{t}\right), \quad
    \left[\hat{\V{e}}^i_{t}, \V{w}^i_{t} \right] = \txt{FC}\left(\V{h}^i_{t,tr}\right)\label{eq:ew}\\
    & \V{H}^{i+1}_{t,y} = \left( \V{1}- \V{A}^i_{t}\otimes \V{e}^i_{t}\right)\otimes \V{H}^i_{t,y} + \V{A}^i_{t}\otimes \V{w}^i_{t}.\label{eq:write}
\end{align}
Here, $\V{e}^i_{t}$ and $\V{w}^i_{t}$ are the erasing and writing vector, respectively. The feature map is upadated in the spatial locations indicated by the attention matrix $\V{A}^i_{t}$ of the previous iteration. The operator $\otimes$ denotes a element-wise product in the channel dimension of $\V{H}^i_{t,y}$. 

\paragraph{Koopman Embedding}
We employ Koopman theory as an alternative to modelling dynamics. We argue that this will introduce benefits to our model, as we discuss in Section \ref{sec:exp}.\\
Assumptions are made with respect to the objects' dynamics in the scene. 
We define the state as a concatenation of $T_{S}$ delayed instances (from now on referred to as delayed coordinates) of the pose vector $\V{z}_{p}^i$ for the $i_\text{th}$ object:
\begin{align}
    &\V{s}_t^i = \left[\V{z}_{t,p}^i, \V{z}_{t-1,p}^i, \dots, \V{z}_{t-T_{S}+1,p}^i \right]
\end{align}
Therefore $\V{s}_t^i \in \mathbb{R}^{4T_s}$. 
$T_s$ indicates our prior belief on the number of time-steps needed to model dynamics with an Auto-Regressive (AR) approach. We assume that the effects of the environment on each object will remain unchanged through training and testing cases. Hence, these effects are learned implicitly by the model. 
The observables $\V{g}_t^i$ are obtained through the mapping $f_{\mathcal{M}}: \Re^{4 \times T_s} \rightarrow \Re^\mathcal{M}$ (equivalent to Eq. \ref{eq:mapping}). $\mathcal{M}$ and $\mathcal{U}$ make reference to the spaces of observables and inputs respectively. We recover the original state by approximating the inverse function $f_{\mathcal{M}^{-1}} \approx f_{\mathcal{M}}^{-1}$ with a deterministic Auto-Encoder (AE) architecture. In the presence of external forces, we will also model inputs $\V{u}_{t}$ as a non-linear mapping $f_\mathcal{U}: \Re^{4T_s} \rightarrow (-1,1)^\mathcal{U}$ from the state space. Note that we set $f_\mathcal{U}: \Re^{4  T_s} \rightarrow \{0\}^\mathcal{U}$ if we assume that the objects' dynamics are not affected by the environment:
\begin{align}
    \hat{\V{s}}_{t+1}^i &=  f_{\mathcal{M}^{-1}} \circ \left( \mathcal{K} \circ f_{\mathcal{M}}\left( \V{s}_{t}^i\right) + \mathcal{K_\mathcal{U}} \circ f_{\mathcal{U}}\left( \V{s}_{t}^i\right)\right) \\ &=  f_{\mathcal{M}^{-1}} \left( \mathcal{K}  \V{g}_{t}^i + \mathcal{K_\mathcal{U}}  \V{u}_{t}^i\right).\label{eq:koopwithu}
\end{align}
We refer to the estimated states as $\hat{\V{s}}_T^i$. 
We define the Koopman operators $\mathcal{K}: \Re^\mathcal{M} \rightarrow \Re^\mathcal{M} $ and $\mathcal{K_\mathcal{U}}: \Re^\mathcal{U} \rightarrow \Re^\mathcal{M}$ as parameter matrices.
The pose vector $\hat{\V{z}_{t+1,p}^i}$ is recovered by keeping the first stacked coordinate of the estimated state $\hat{\V{s}}_{t+1}^i$, and limited to the range $(-1, 1)$.
The eigendecomposition of the Koopman operator $\mathcal{K}$ provides us with insights of the dynamics in the scene. 

\paragraph{Training and Objective}

Given the video sequence $(\V{y}_1,\cdots, \V{y}_T)$, its generative distribution is given by:
%
\begin{eqnarray}
    p(\V{y}_{1:T}|\V{z}_{1:T})= \prod_{i=1}^N p(\V{y}^i_{1:T}|\V{z}^i_{1:T})p( \V{y}^i_{1:T}|\V{z}^i_{1:K}),
    \label{eq:gen}
\end{eqnarray}
%
where $\V{z}^i_{t} = \left[z^i_{t,c}, \V{z}^i_{t,p}, \V{z}^i_{t,a} \right]$. Note that we both reconstruct the whole sequence with length $T$ and predict it from $K$ initial frames $(K<T)$. For our experiments, $K$ will coincide with the number of delayed coordinates $T_s$.\\
Given the inferred latent variables, we reconstruct and predict  $\V{y}_i^t$ for each  object sequentially. In particular, we first generate the object in the center with resolution $R = C \times h \times w$, given the appearance $\V{z}^i_{t,a}$. The decoder $f_{\txt{dec}}: \Re^A \rightarrow \Re^R$ is a deconvolutional layer. 
We then apply a spatial transformer $\T{T}^{}$ to rescale and place the object according to the pose $\V{z}^{i}_{t,p}$. For each object, the generative model is:
\begin{eqnarray}
p(\V{y}^i_t|\V{z}^i_{t,a}) = \T{T}(f_{\txt{dec}}(\V{z}^i_{t,a}); \V{z}^{i}_{t,p})\circ  z_{t,c}^i, 
\label{eq:gen_model}
\end{eqnarray}
Future prediction is similar to  reconstruction, except in this case $\hat{\V{z}}^{i}_{t,p}$ is extrapolated using the Koopman operator in the observable space. \\
The generated frame $\V{y}_t$ is the summation over $\V{y}_i^t$ for all objects.
Similarly to the VAE framework, we train the model by maximizing the evidence lower bound (ELBO). 
\begin{equation}
\begin{split}
    &\log p_{\theta}({\V{y}}_{1:T}) \ge 
    \mathbb{E}_q\big[\log p_{\theta}\left({\V{y}}_{1:T}|{\V{z}}_{1:T}^{1:N}\right) + \\
    &\log p_{\theta}\left({\V{y}}_{K+1:T}|{\hat{\V{z}}}_{K+1:T|1:K}^{1:N}\right) - \\
    &\text{KL}\left(q_{\phi}\left({\V{z}}_{1:K,a}^{1:N} \right) ||p\left({\V{z}}_{1:K,a}^{1:N}\right)\right) \big]
\end{split}
\end{equation}
Here, we use self-supervision for reconstructing the input $\V{y}_{1:T}$ and predicting that same input from few initial conditions ($1:K$). 
We also add regularizers for a better learning of the Koopman embedding, so that the final expression of our objective $\mathcal{J}_{\omega}$ with respect to the trainable weights $\omega$ is:
\begin{align}
    &\mathcal{J}_{\omega} = 
    \min_{\omega}{\left[-\text{ELBO}\! +\! L_{\txt{fit}}\! + \!L_{\txt{AE}}\! + \! L_{\mathcal{U}}\! +\! L_{\txt{Rank}\mathcal{K}}\right]}
    \\
    &L_{\txt{AE}}\! = \! \!\lambda_{\txt{AE}}\!\left( \left\| \V{s}_{3:T}^{1:N}\! -\! \hat{\V{s}}_{3:T | 1:T\!-\!2}^{1:N}\right\|_1
    \!\!+\!\! \left\| \V{s}_{T_s:T}^{1:N}\! -\! \hat{\V{s}}_{T_s\!+\!1:T | T_s}^{1:N}\right\|_1\!\right)\!\nonumber
    \\
     &L_{\txt{fit}} = \lambda_{\txt{fit}}\left\| \V{g}_{2:T}^{1:N} - \hat{\V{g}}_{2:T|1:T-1}^{1:N}\right\|^2_2 ,\nonumber\\
     &L_{\mathcal{U}} = \lambda_{\mathcal{U}} \left\| \hat{\V{u}_{1:T}^{1:N}}\right\|_{1}, \quad
     L_{\txt{Rank}\mathcal{K}} =\lambda_{\txt{r}\mathcal{K}} \left\| \mathcal{K}\right\|_{*}\nonumber
\end{align}
Here, $\left\|\cdot \right\|_1$ and $\left\|\cdot \right\|^2_2$ indicate the $\ell_1$ and $\ell_2$ losses respectively. $\left\| \cdot \right\|^*$ is the nuclear norm of a matrix, which is a convex surrogate of the rank function. $\left\|\cdot \right\|_1$ is used to enforce sparsity in $\hat{\V{u}}_{1:T}^{1:N}$. Finally, the $\lambda$s are the weights applied to each one of the loss particles.
In terms of implementation, we use linear annealing to increase linearly the weights $\lambda$ of the regularizers as training advances. For more details see the Appendix.

\section{Experiments}\label{sec:exp}
We evaluate \ours{} in variations of the Moving MNIST dataset. Our goal is to show that our method can capture and model the implicit dynamics of objects in video sequences with a Koopman embedding. Therefore, we will preserve the same distribution of appearances through the experiments (MNIST digits), and vary the nature of their trajectories. Our baselines are the established state-of-the-art methods for decomposed self-supervised video generation: \textsc{ddpae} \cite{ddpae}, \textsc{drnet} \cite{drnet} and \textsc{scalor} \cite{scalor} 
All of them base their dynamics modelling on RNNs. 

\paragraph{Evaluation Metrics} Our quantitative results will be measured in terms of pixel-level Binary Cross entropy (BCE) per frame, Mean Square Error (MSE) per frame, Mean Absolute Error (MAE), Structural Similarity (SSIM) and the Perceptual Similarity Metric (LPIPS) \cite{lpips}.

\subsection{Implementation}
Our model is trained with different configurations for every experiment. More details are provided in the Appendix. When it comes to the dimensionality of the observable space $\V{g}_t^i: \Re^\mathcal{M}$, it ranges from 15 to 30. We use an appearance vector of size 50, a pose vector of size 4 and the number of delayed coordinates per state $\V{s}_t^i$, $T_s$, corresponds to the number of input frames in our experiments. The tracker hidden state corresponding to Equation \ref{eq:tracker_state} has dimension 288. The tracker attends to a $96\times4\times4$ convolutional map.
The objects are decoded to a size of $1 \times 32 \times 32$ and located in the $64 \times 64$ frame.

\subsection{Moving MNIST Experiments}
Moving MNIST \cite{Srivastava2015UnsupervisedLO} is a synthetic dataset consisting of two digits with size $28\times28$ moving independently in a $64\times64$ frame. Each sequence is generated on-the-fly by sampling MNIST digits and synthesizing trajectories according to a definition of the motion. Our model is trained for 200 epochs. We randomly generate 9k sequences for training, 1k for validation and 2k for testing. In our experiments, we simulate 4 scenarios as follows:
\begin{itemize}
    \item \textbf{\textit{Circular motion}}: For the first experiment, we generate a fairly simple dataset, for which we know the expected results. We sample randomly initial coordinates $(x_0,y_0)$, radius $R$ and the angular step length. We generate the motion with equation: $x = R\cos{(t)}+x_0$, $y=R\sin{(t)} + y_0$, where $t$ increases linearly with a slope given by the angular step length. Finally, we constrain the motion to the dimensionality of the frame. We generate $T=20$ frames as our input. From the first 3, we will generate 17 with supervision.
    \item \textbf{\textit{Cropped circular motion}}: We mask the 29 top rows of the circular motion case, simulating a partially cropped frame. Note that an object of size $28\times28$ can be completely occluded. 
    \item \textbf{\textit{Inelastic/Superelastic collisions}}: With a fixed velocity, we sample initial coordinates $(x_0,y_0)$ and angle $\theta$ and let the object collide against the frame limits. We increase the complexity of the case by simulating an inelastic response from the left and top limits $(\times 0.8)$ and a superelastic response from the right and bottom limits $(\times 1.25)$. We generate chunks of $T=13$ frames as input. From the first 3, we will generate 10 with supervision.
    \item \textbf{\textit{3D to 2D motion projection}}: This motion is created parametrically following $z =cos(v_z\theta+\phi_{z_0})$, $R = \sqrt{1-z^2}$, $x = R\cos{(v_x\theta+\phi_{x_0})}$, $y = R\sin{(v_y\theta)}$, with coordinates $(x,y,z)$ and angular velocities $(v_x, v_y, v_z)$.
    This parameterization constrains it to lay within the cube $[-1, 1]^3$. We then rotate the trajectory 
    with an angle $\pi/8$ and project a random portion of the full trajectory with size $T=16$ (6 in 10 out) to the $xy-$ axis. The offset phase $\phi_{z_0}$ and $\phi_{x_0}$ are also randomly generated. Using perspective projection, the objects are resized according to their depth in the $z$ axis, after being projected. Figure \ref{fig:figure_sph} gives an example of the generated sequences.
    \begin{figure}[h]
    \centering
    \includegraphics[width=0.25\textwidth]{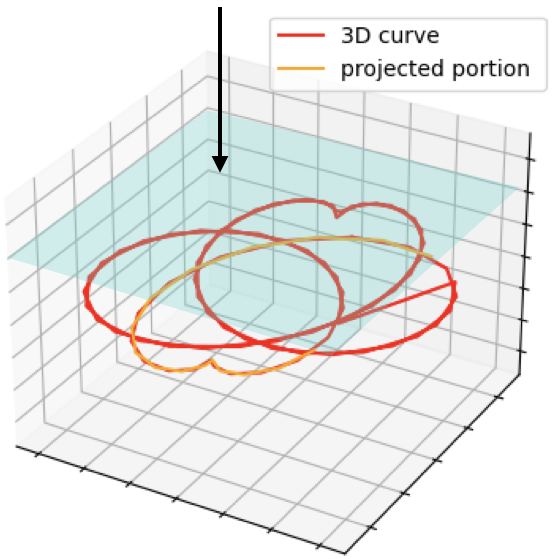}
    \caption{Example of a projected trajectory for \textit{3D to 2D motion projection} experiment. In orange the selected chunck.
    }\label{fig:figure_sph}
\end{figure}  
\end{itemize}
The number of input frames will contain enough information to predict auto-regressively in most sequences of each scenario. Therefore, we use $T_s$ delayed coordinates for our state $\V{s}_t$ corresponding to the input frames.
More details on the dataset can be found in the Appendix.\\

\begin{figure}[hbpt!]
\centering
\includegraphics[height=0.2\textwidth]{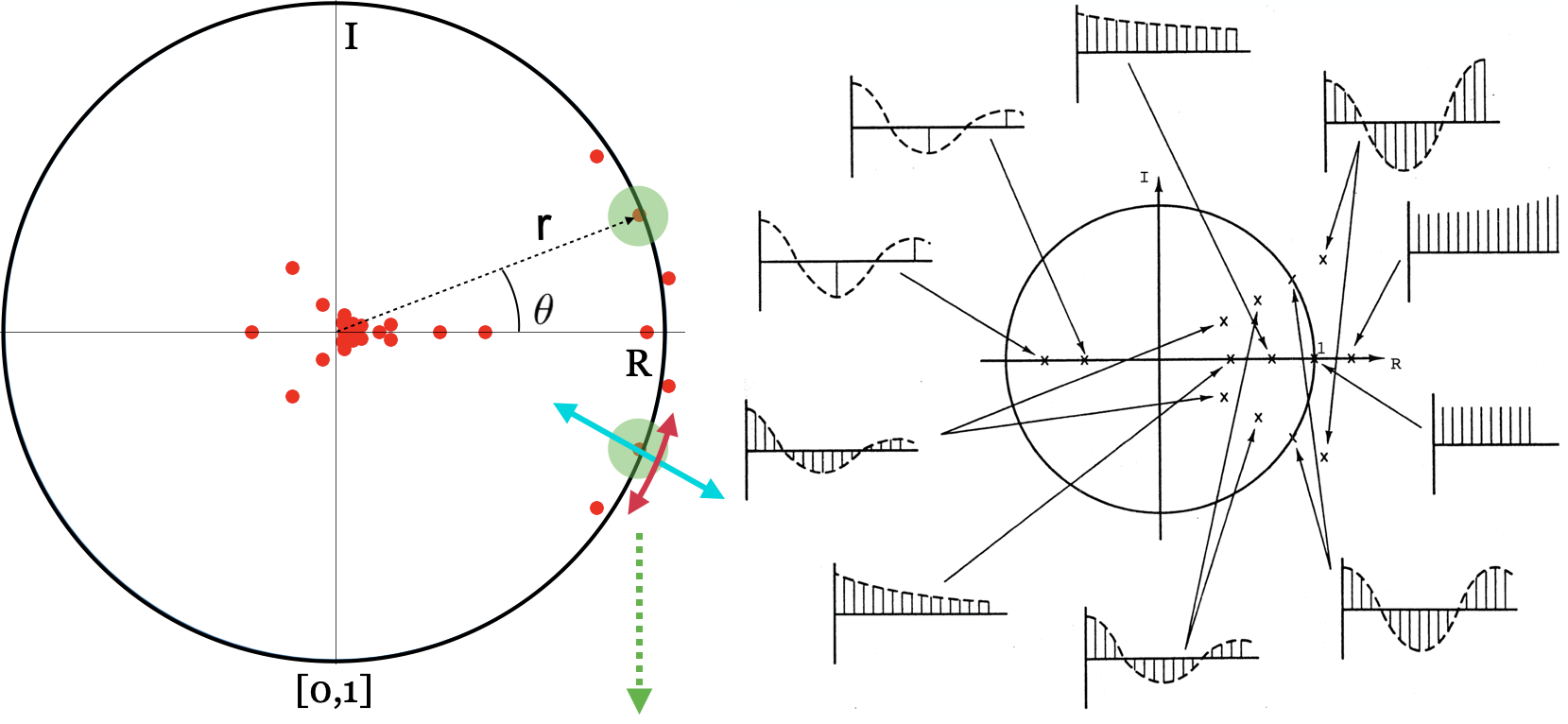}\\
\vspace{2mm}
\includegraphics[height=0.77\textwidth]{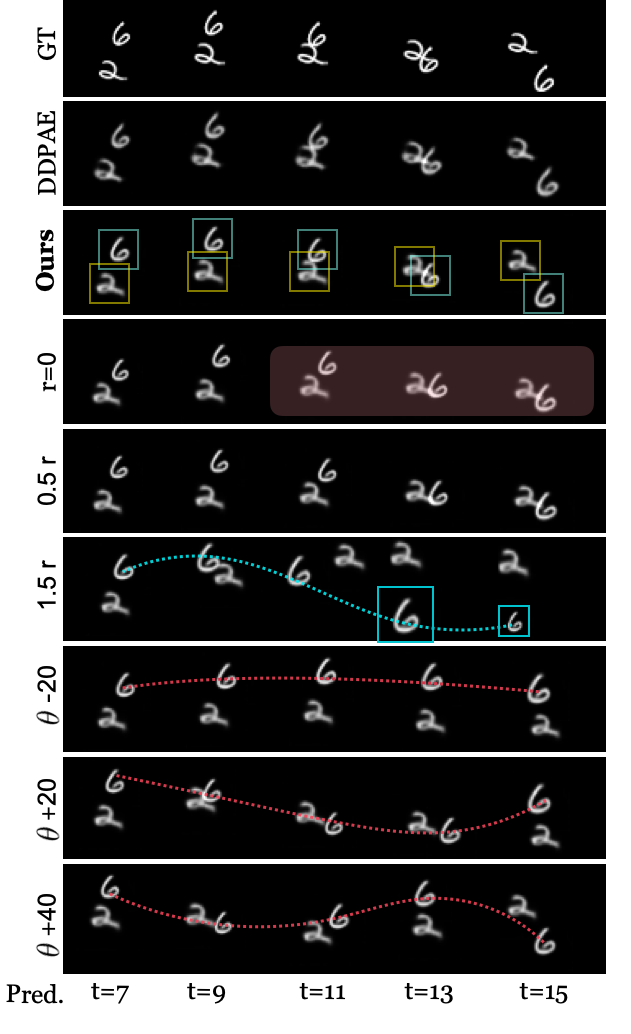}
\vspace{-2mm}
\caption{Top right: Visualization of the atomic dynamics given by the eigenvalues of matrix $\mathcal{K}$ \cite{phillips2007digital}. Top left: Learned Koopman matrix eigenvalues for 3D to 2D projection experiment. In green we highlight the eigenvalue pair that we will modify to visualize the effect of manipulations. Bottom, predictions by rows: Ground Truth; our main baseline \textsc{ddpae}; \ours{} with the object decomposition bounding boxes; 2 variations to the radius of the highlighted eigenvalue; 3 variations to the angle of the eigenvalue. We can see in blue, how increasing the radius increases the effect of that particular eigenvalue, both in size and position. In red, a modification in the angle has a direct effect on the frequency of the motion.}\label{fig:results_sph}
\end{figure}

A quantitative general overview of the experiments can be seen in Table \ref{tab:results}. \ours{} outperforms the baselines in most cases except for \textsc{scalor} in reconstruction, and \textsc{ddpae}. In terms of Perceptual Similarity (LPIPS), our prediction outperforms all the baselines. \textsc{ddpae} is the closest to ours in terms of architecture. The key difference is the dynamics modeling. \textsc{ddpae} uses a concatenation of LSTMs for reconstructing and predicting the pose. Each of them has a hidden state of size $64$. 
Our objective is not to outperform RNN-based methods for short sequences such as video. We aim to perform similarly while gaining interpretability and manipulability of the dynamics. 
In the same table, we can see the results of \ours{} after keeping the top $6$ eigenvalues of the Koopman matrix $\mathcal{K}$, what is known as model reduction. The actual dimensionality of $\mathcal{K}$ ranges from $15 \times 15$ to $30 \times 30$ in the experiments. As we see, the results are very similar to the prediction with the full matrix. This means that our model was able to capture the motion of the digits with only 3 conjugate pairs of eigenvalues in all studied cases.
Figure \ref{fig:results_sph} illustrates how single or conjugated pairs of eigenvalues impact on the dynamics. The produced dynamics will be a weighted combination of the impulse responses produced each eigenvalue (real or conjuagate pair).
In the same figure, we see a manipulation of the learned model for the \textit{3D to 2D motion projection} case. Qualitatively, the generated frames are sharp and accurate, and the learned dynamics are correct. Note that even after modifying the eigenvalues of matrix $\mathcal{K}$, the motion of the digits is physically plausible and smooth.

\begin{figure}[t]
\centering
\includegraphics[height=0.25\textwidth]{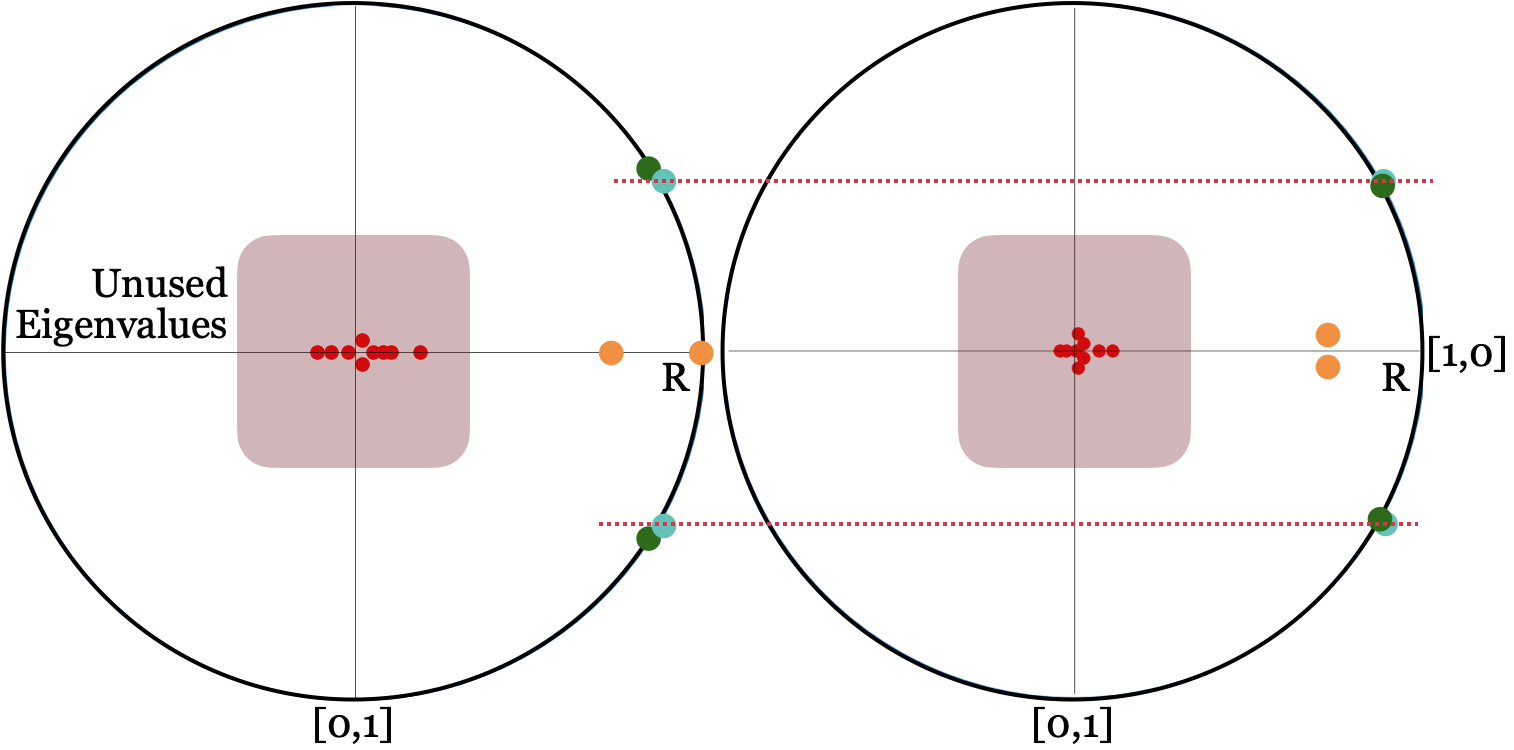}
\vspace{2mm}
\caption{Visualization of the eigenvalues of the learned matrix $\mathcal{K}$ for both \textit{circular} experiments. In the left, for the complete frame experiment; in the right, for the cropped frame experiment. We can see how the learned eigenvalues are very close in both cases, while one of them has been trained in a corrupted dataset. In red, we highlight the eigenvalues of $\mathcal{K}$ that have been empirically shown to have almost no effect on the dynamics modeling.}\label{fig:eigenvalues_circ}
\label{fig:atoms}
\end{figure}

\paragraph{Circular Motion Experiments}
Table \ref{tab:results} shows the results for this scenario. We see that we perform similarly to \textsc{ddpae} in both the cropped and the complete version of the experiment. Here, we know that the motion is expected to be sinusoidal in both $x$ and $y$. Therefore, a correct Koopman mapping would not need to be non-linear to capture the dynamics in this case. A sinusoid can be modelled by a linear operator with a complex pair of eigenvectors in the unit circle. In Figure \ref{fig:eigenvalues_circ}, we can see that pair, together with real eigenvalues that present no oscillation. We observe also that the learned model is stable as no eigenvalue has greater radius than 1. The model has learned a very similar operator $\mathcal{K}$ for both the cropped and the complete version. This is an indicator that \ours{} is learning consistent dynamics across datasets when they share the same motion. It also suggests that the model is able to impute a trajectory when the data is missing. Qualitative results can be seen in the Appendix.

\paragraph{Collision Experiments} For this scenario, the challenge is the use of inputs $\V{u}_t$. Every collision against the frame limits applies a force to the object, that modifies its dynamics. Therefore, we model the effect of the environment as in Equation \ref{eq:koopwithu}, allowing the inputs to be non-zero, and forcing them to be sparse and low-dimensinoal $(\V{u}_t \in \Re^4)$ to avoid overfitting. This generates sharp objects and captures correctly the dynamics. See Appendix for more details.

\paragraph{3D to 2D Projection Experiments}. 
Quantitative results for this experiment are fairly close to \textsc{ddpae}, especially when it comes to prediction. This dataset is challenging because it entangles linear motion across dimensions by means of projection. It also encompasses digit size variations. However \ours{} is able to estimate the dynamics correctly in most cases, and generates sharp-looking objects. Figure \ref{fig:results_sph} shows the qualitative performance in terms of prediction of \ours{}. We can see that the predictions are accurate and sharp. The model correctly disentangles the two objects that appear in the scene, and models their dynamics independently. To understand and interpret the dynamics learned in the Koopman space, we modify the the eigenvalues of the operator $\mathcal{K}$. As shown in Fig. \ref{fig:results_sph} top, we modify the eigenvalues highlighted in green by changing their radius $r$ or their angle $\theta$. We can interpret the following from the \ref{fig:results_sph} bottom: Shadowed in red, we can see that this particular eigenvalue pair has effect in the latter part of the trajectory. If we increase its module above 1, we observe an increase on the intesity of the variations, that seem to oscilate strongly at the end of the sequence (see size of the object). This happens because the system is now unstable. If we vary the angle of the eigenvalue pair with respect to the real axis, we see variations in terms of frequency. When we subtract 20 degrees to the angle, it is almost 0. Therefore, we observe a constant trajectory for the objects. When we increase that angle, we see the frequency of the digits oscillation increasing with it. 

This is a clear example of how Koopman allows us to interpret and manipulate the modelled dynamics.

\begin{table*}[htbp!]
  \caption{Quantities comparison of all methods for the four scenarios. We evaluate reconstruction, prediction, and prediction with model reduction (6e denotes that we keep only the top 6 eigenvalues). From top to bottom: Circular motion, Cropped circular motion, Inelastic/Superelastic collisions, 3D to 2D motion projection. Our method performs similarly to the best RNN-based baseline and outperforms the rest of baselines in prediction.}\label{tab:results}
  \small
  \centering
  \setlength\tabcolsep{5pt}
  \resizebox{1\textwidth}{!}{
  \begin{tabular}{c|ccc|ccc|ccc|ccc|ccc}
    \toprule
        Model& \multicolumn{3}{|c|}{MSE $\downarrow$}
            & \multicolumn{3}{|c|}{MAE $\downarrow$}
                & \multicolumn{3}{|c|}{BCE $\downarrow$}
                    & \multicolumn{3}{|c|}{SSIM $\uparrow$}
                        &\multicolumn{3}{|c|}{LPIPS $\downarrow$}\\
    \toprule
    \toprule
   Circular &Rec&Pred&pred(6e)&Rec&Pred&pred(6e)&Rec&Pred&pred(6e)&Rec&Pred&pred(6e)&Rec&Pred&pred(6e)\\
    \hline
    \midrule
    
    \textsc{ddpae}\cite{ddpae}   & 40.13 & 71.96 & / &  123.94 &162.64& / & 225.67 & 331.80 & /& 0.87& 0.82& / & 0.19 & 0.21 & / \\
    \textbf{\textsc{\ours{}}}  &   59.97     &  84.23 &84.64& 139.92&168.83&169.96& 283.26& 371.46& 317.80& 0.86 & 0.82 & 0.82 & 0.15 & 0.17 &0.17\\
    \toprule
     Cropped circular &Rec&Pred&pred(6e)&Rec&Pred&pred(6e)&Rec&Pred&pred(6e)&Rec&Pred&pred(6e)&Rec&Pred&pred(6e)\\
           \hline
      \midrule
    \textsc{ddpae}\cite{ddpae}&35.04 & 54.95 & / &98.48&118.90&/& 176.09& 248.25&/ & 0.88&0.85&/&0.21&0.23&/ \\
    \textbf{\textsc{\ours{}}}  &   47.80 &  59.26  & 59.32  & 108.28 & 120.04& 120.15& 212.84 & 254.09 & 254.24 & 0.88 & 0.86 & 0.86 & 0.17 & 0.19 & 0.19\\
    \hline
    \toprule
    Collision &Rec&Pred&pred(6e)&Rec&Pred&pred(6e)&Rec&Pred&pred(6e)&Rec&Pred&pred(6e)&Rec&Pred&pred(6e)\\
    \hline
    \midrule
    \textsc{drnet}\cite{drnet}   & 109.01 & 214.49 &/&  218.14& 339.58 &/&478.39&1350.20&/&0.75&0.60&/&0.30&0.40&/  \\
    \textsc{scalor}\cite{scalor}   &   13.24 &329.63 &/& 50.02& 494.41&/ &148.02&1584.48&/&0.95&0.28&/&0.02&0.48&/ \\
    \textsc{ddpae}\cite{ddpae}  &   49.53 & 93.52  &/ & 146.65 &199.57&/ &263.39& 423.88&/&0.84&0.76&/&0.23&0.25&/ \\
    \textbf{\textsc{\ours{}}}  &   59.53     & 103.63 &110
    & 155.07 &205.31&212.38&291.18&449.73&473.68&0.83&0.77&0.76&0.19&0.21&0.22\\
    \bottomrule

    \toprule
    3D to 2D proj &Rec&Pred&pred(6e)&Rec&Pred&pred(6e)&Rec&Pred&pred(6e)&Rec&Pred&pred(6e)&Rec&Pred&pred(6e)\\
    \hline
    \midrule
    \textsc{drnet}\cite{drnet}    & 80.31 & 136.06&/ & 172.43& 248.11&/& 331.71&732.60&/&0.77&0.66&/ &0.32&0.41&/\\
    \textsc{scalor}\cite{scalor}   &  7.58 & 233.18  &/& 32.05& 377.34 &/&87.95&1018.30&/&0.96&0.32&/& 0.02&0.45&/ \\
    \textsc{ddpae} \cite{ddpae} &   20.99 & 43.72  &/ & 57.08 &86.87&/ &124.45& 223.50&/&0.94&0.89&/&0.11&0.14&/ \\
    \textbf{\textsc{\ours{}}}  &   38.15     &  45.22 & 48.25 & 85.37 &92.67&96.27 & 188.46 &216  &228.97&0.89&0.88&0.87&0.15&0.14&0.15\\
    \bottomrule
    \hline
    
  \end{tabular}
  }
  \vspace{5mm}
\end{table*}

\newpage

\section{Conclusion}
We propose a self-supervised method for decomposing a video sequence into interpretable components. In the spirit of compositionality, we decompose a scene into its composing objects and the objects into their attributes of pose, appearance and confidence. We leverage the dynamic components to learn a Koopman-based model for their dynamics. We embed the dynamic attributes of each object into a space where dynamics are linear, and therefore prediction is performed by a linear operator $\mathcal{K}: \Re^\mathcal{M} \rightarrow \Re^\mathcal{M}$ and the sparse influence of inputs from the environment. This design enables us to completely decompose a video into meaningful and interpretable components, including its evolution in time.
The key advantage of our model is that we can utilize tools in control theory to understand the underlying dynamical system in a high-dimensional and highly nonlinear sequence such as video; for instance, we provided insights into the dynamics of objects through the analysis of eigenvalues of learned Koopman operator. Through carefully designed experiments, we showed that our method does not sacrifice accuracy or predictability while maintaining interpretability.
To the best of our knowledge, our work is the first to introduce Koopman analysis in the interpretation of video sequences. We are excited that this opens the door to exchange ideas between computer vision, control system and interpretability, thereby allowing the theoretical development and analysis in the area of control systems to positively impact interpretation of video sequences.
In the future, we wish to extend our method to handle more complex videos involving interactions between objects and between objects and the environment simultaneously. We also intend to extend this implementation into handling the background in parallel to the foreground. This would allow \ours{} to model more complex environments.

\newpage
{\small
\bibliographystyle{ieee_fullname}
\bibliography{kidd}
}

\clearpage

\normalsize

\appendix
For a better understanding of this work, please see the slideshow given as part of the supplemental material. 

\setcounter{figure}{0}  
\counterwithin{figure}{section}

\section{Model Implementation Details}
\label{app:model_details}

\begin{figure*}[t]
\centering
\includegraphics[height=0.4\textwidth]{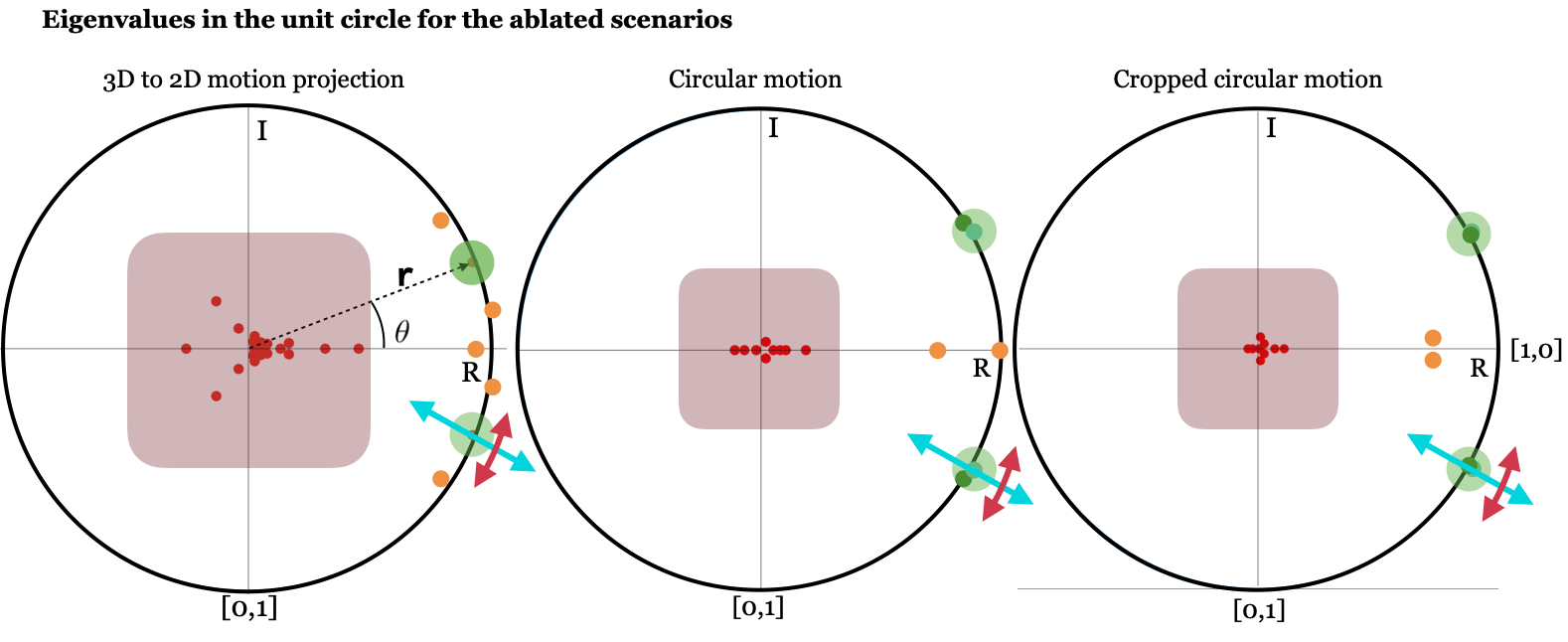}\\
\caption{Eigenvalues in the unit circle for the Koopman operator $\mathcal{K}$ for the three ablated scenarios. The axes are the imaginary (I) and real (R) components of the eigenvalues. Shadowed in red, the area where the eigenvalues have negligible effect on the trajectories, according to the results shown in Table 1.
The following indicators are linked to the results shown in Figures \ref{fig:sph-app}, \ref{fig:circ-app}, and \ref{fig:crop-app}: Highlighted in green, the chosen eigenvalue pair for the ablation study. The blue and red arrows indicate the kind of manipulation that the chosen eigenvalue pair will undergo. $r$ and $\theta$ are the magnitude and phase of the uppermost eigenvalue with respect to the real axis.}\label{fig:eig-app}
\end{figure*}

\subsection{Baseline Models}

\paragraph{\textsc{scalor}} The \textsc{scalor}\cite{scalor} model  is an unsupervised model for learning scalable object oriented representations. The model can track a large amount of objects with a dynamic background. However, although \textsc{scalor} can predcit future frames from previous frames, its model is not specifically designed for  prediction, although prediction is reported. The baseline was chosen as an alternative for parallel object decomposition in an attention-based tracking. In this work, our experiments show that it \textsc{scalor} is very good at reconstruction, but predicts very poorly. Thus, we were not able to reach a reasonable performance in terms of prediction for the studied dynamical scenes. 

For our experiments we used the implementation available in \url{https://github.com/JindongJiang/SCALOR.git}. In order to obtain the best performance with \textsc{scalor} on our data set, we trained the model for 625 epochs. However, we found that the loss reached a plateau for the prediction task in the early stages. At that point of training, \textsc{scalor} is unable to generate predictions. In order to tackle this phenomenon, we kept the MNIST digits stationary and trained \textsc{scalor} until it could generate predictions. From that point, we observed that after 625 epochs of training, we got some good prediction frames. Nevertheless, predictions are often blurry and inaccurate, and obvious artifacts appear in the scene. The training time per 100 epochs for \textsc{scalor} is around 30 minutes with 4 RTX 2080 Ti GPU with 12.8GB memory  each of them.

\paragraph{\textsc{drnet}} The original version  of the \textsc{drnet}\cite{drnet} model only uses the first four frames as input for training. For our experiments, we need the three/six input frames. We changed the scene discriminator in \textsc{drnet} to train on all frames in the sequences. The rest of the model was kept exactly the same as the authors' implementation for better reproduction of results. As mentioned in their github repository (\url{https://github.com/ap229997/DRNET.git}), the main network and the \textsc{lstm} in \textsc{drnet} were trained separately. Firstly, we trained the main base network, then we trained it again with skip connection, and finally we trained the \textsc{lstm} part. Therefore, the performance of the the \textsc{lstm} part is determined by the main network. The scene discriminator was trained with BCE loss. The main network and \textsc{lstm} were trained with MSE loss; and we trained the main model and \textsc{lstm} with 4 RTX 2080 Ti GPU with 12.8GB memory each. For more details about \textsc{drnet}, please refer to their github.

\paragraph{\textsc{ddpae}} We used the code provided by the authors. The hyperparameters that they use in the public version were kept unchanged. Also, we followed the instructions in their github repository (\url{https://github.com/jthsieh/DDPAE-video-prediction}) for the Moving MNIST experiment. The model was trained for 250 epochs, similarly to \ours{}.


\subsection{Our Model}

\paragraph{\ours} The main latent variables have the following dimensions: $\V{z}_{t,a}^i \in \mathbb{R}^{50}$, $\V{z}_{t,c}^i \in \left[0,1\right)$ and $\V{z}_{t,p}^i \in \mathbb{R}^4$. The latter's  four dimensions  correspond to $\left[x, y, \Delta s, \Delta r \right]$, where $x$ and $y$ are the coordinates of the centroid of an object; $\Delta s$ is the increment of the size of the object with respect to the decoded $32 \times 32$ appearance, which would be of size $s=1$; and $\Delta r$ is the increment of the ratio $s_x/s_y$, which by default is $1$. The increments are weighted by a scalar $w \in (0, 1]$ that regulates their effect. All components of the pose are bounded between $-1$ and $1$. 

Following the VAE framework, we implemented $\V{z}_{t,a}^i$ to be a sample of a learned posterior distribution $q\left(\V{z_{t,a}^i|\V{y_t}}\right) = \mathbb{N}(\mu_a, \sigma_a)$, with the reparametrization trick. As usual, we regularized our training by adding a KL divergence term between our posterior and a Gaussian prior with $\mu = 0$ and $\sigma = 1$.

We implemented the convolutional features with dimensionality $\V{H}_{t,y}^i \in \mathbb{R}^{(96 \times 4 \times 4)}$ and the tracker hidden states as $\V{h}_{t,tr}^i \in \mathbb{R}^{288}$. 
The dimensions were chosen after a manual sweep of hyperparameters range. Particularly, the dimensionality of $\V{h}_{t,tr}^i$ was chosen from the range $[96, 192, 288, 384]$;  $\V{z}_{t,a}^i$ from $[40, 50, 70]$; and $\V{H}_{t,y}^i$ from $[96, 128]$. The Koopman mapping and inverse is parametrized by a multi-layer perceptron with 4 layers and hidden state dimensionality of 40. The Koopman operator $\mathcal{K}$ is initialized as a matrix of 0s, and the input operator $\mathcal{K}_\mathcal{U}$ with Xavier initialization (same as other layers of the Koopman embedding).
$\V{H}_{t,y}^i$ was obtained by leveraging the input frames $\V{y}_t$ and a positional encoding $\textsc{PE}$. The latter has dimensionality $4$ and values that indicate the distance from the $4$ frame edges in terms, normalized in $[0,1]$. With the exception of these details, the implementation of the attention and memory follows the major guidelines of \cite{He2019TrackingBA}.

We trained the model in all scenarios for 250 epochs and 9k iterations per epoch. We used a batch size of $40$, and a prior number of objects $N=2$. Similarly to \cite{He2019TrackingBA} or \cite{ddpae}, our model can set redundant components to be empty by reducing the confidence value to 0. The learning rate was set to $5^{-4}$ and reduced by a factor of 0.7 on plateau of the validation loss. We used Adam as our optimizer with parameters $\beta = [0.9, 0.999]$ and weight decay regularization $1e-4$. 


Next, we describe  variations for each experiment.
For the \textit{circular motion} experiment, we have an observable space of dimension $\V{g}_t^i \in \Re^15$ and inputs are set to 0. The same is done for the \textit{cropped circular motion} experiment, but in this case the loss is evaluated only in the visible part of the frame. Note that this is also done  for the baselines. The loss weights are increased linearly, according to the values that will be provided in the codebase.

We also use $\V{g}_t^i \in \Re^15$  for the \textit{Inelastic/Superelastic Collision} experiment. In this case, the input dimensionality is $\V{u}_t^i \in \Re^4$. We keep it low dimensional so it does not absorb the free dynamics of the object.

Finally, for the \textit{3D to 2D motion projection} experiment, we expect higher order dynamics in the Koopman manifold, given the apparent complexity of the dataset. Therefore, we set an observable space of $\V{g}_t^i \in \Re^15$.

Further details can be found in the codebase that will be provided together with the final version of the work.

\paragraph{Software} We implemented this method using Ubuntu 18.04, Python 3.6, Pytorch 1.2.0 and Cuda 10.0.

\paragraph{Hardware} For each of our experiments we used 2 GPUs RTX 2080 Ti (Blower Edition) with 12.8GB of memory.

\begin{figure*}[h]
\centering
\includegraphics[height=0.8\textwidth]{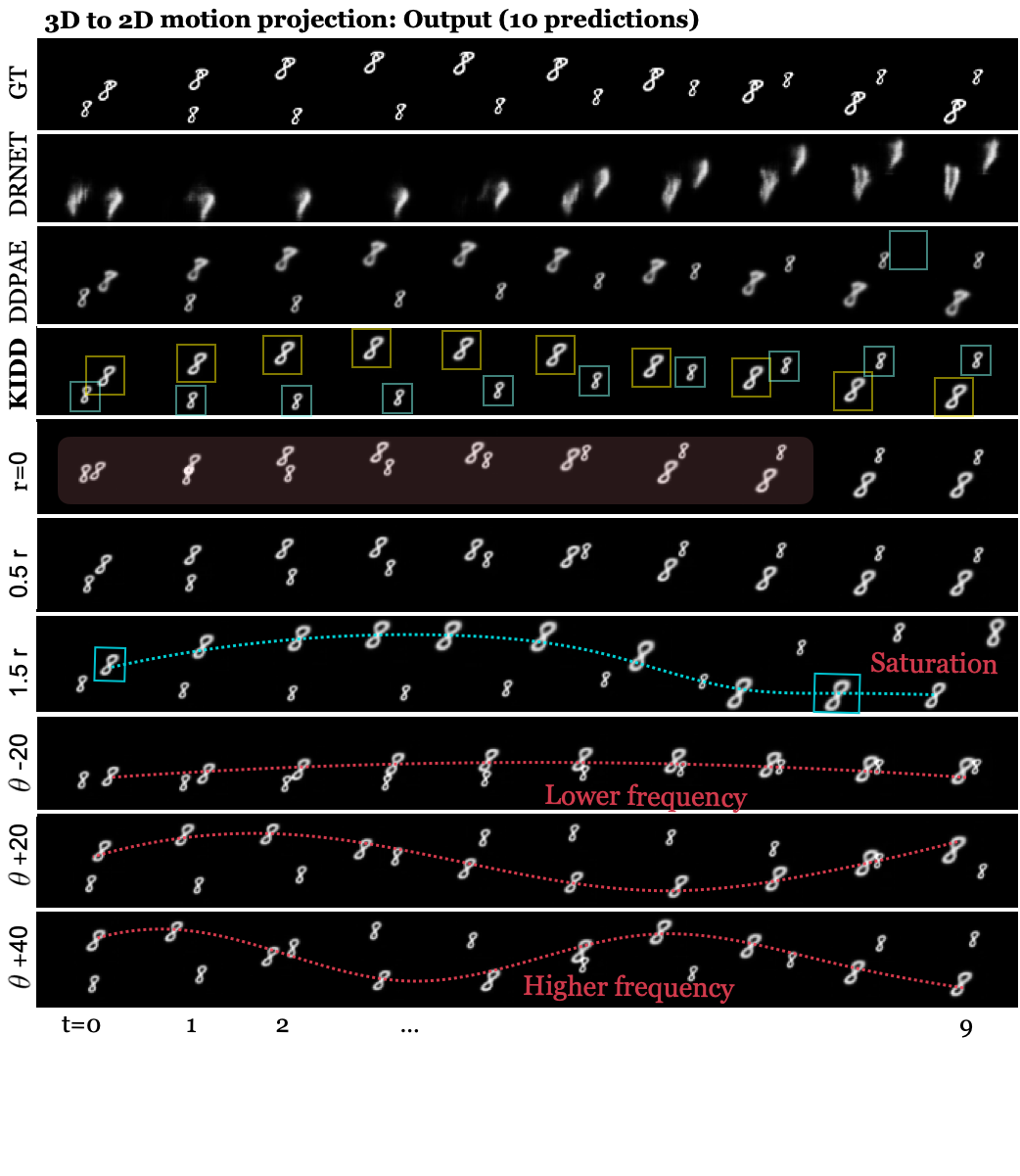}\\
\vspace{-1cm}
\caption{Ablation study and qualitative results of the \textit{3D to 2D motion projection} scenario. The labels in the Y axis indicate the variations sufferd by a selected eigenvalue of the learned $\mathcal{K}$. Written in red, we indicate the behaviors we perceive. We show the $T=10$ predictions of our model and the strongest baseline, \textsc{ddpae}. This setup will be the same as in the rest of ablation studies shown (Figures \ref{fig:circ-app} and \ref{fig:crop-app}). We can see how the model decomposes the scene into its composing objects and predicts accurately their trajectory. We vary one of the eigenvalues (Figure \ref{fig:eig-app}), chosen so that the visualization is clear. We see how variations in angle $\theta$ (red dotted line) have an effect in the frequency of oscillation of the trajectory. Variations in the radius $r$ (blue dotted line) show unstable behaviours when the eigenvalue is greater than 1, and smoothing effects when it's smaller than 1. }\label{fig:sph-app}
\end{figure*}

\begin{figure*}[t!]
\centering
\includegraphics[height=0.8\textwidth]{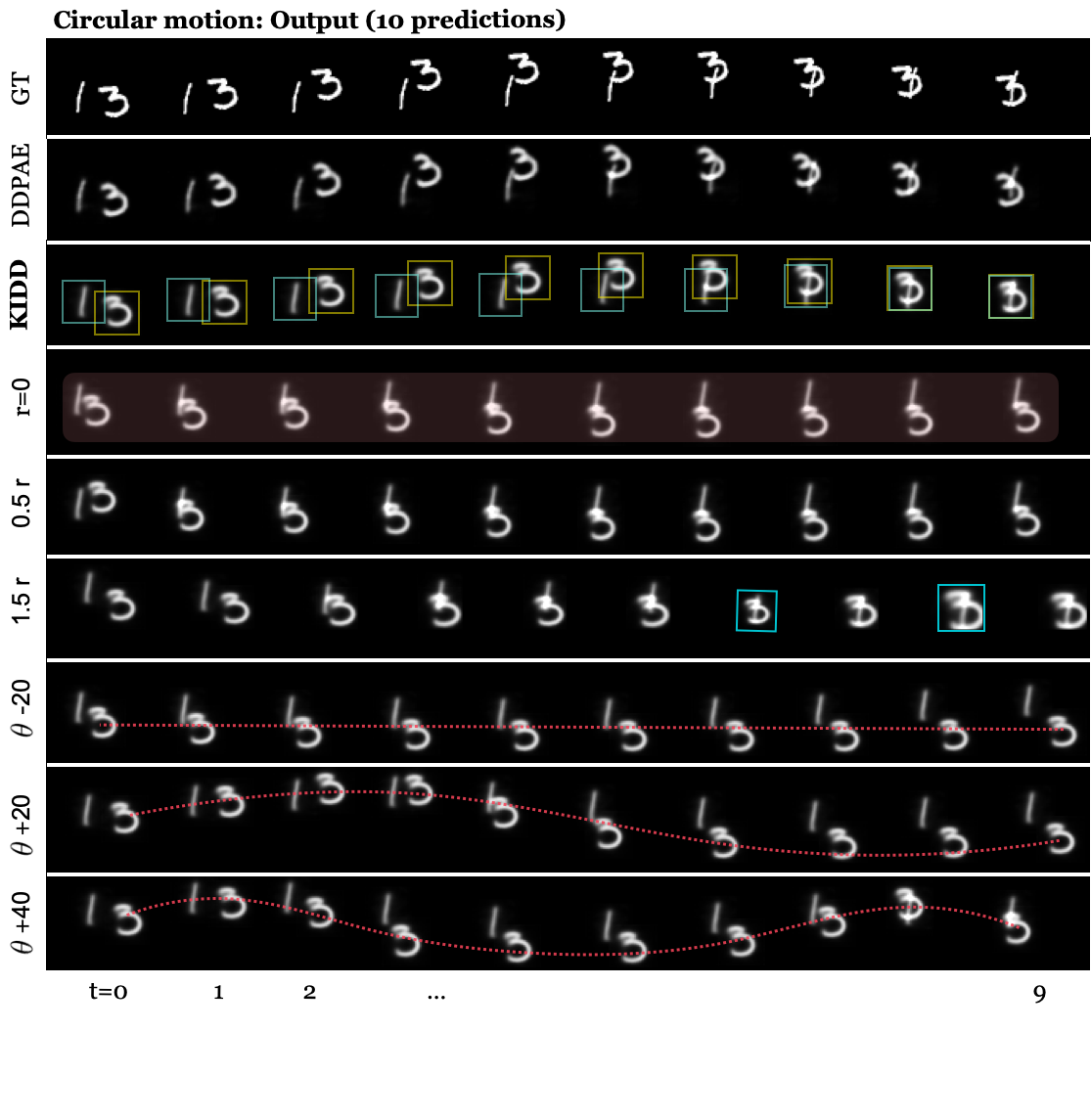}\\
\vspace{-4mm}
\caption{Ablation study and qualitative results of the \textit{Circular motion} scenario. The labels in the Y axis indicate the variations sufferd by a selected eigenvalue of the learned $\mathcal{K}$. Written in red, we indicate the behaviors we perceive.}\label{fig:circ-app}
\end{figure*}

\begin{figure*}[t]
\centering
\includegraphics[height=0.8\textwidth]{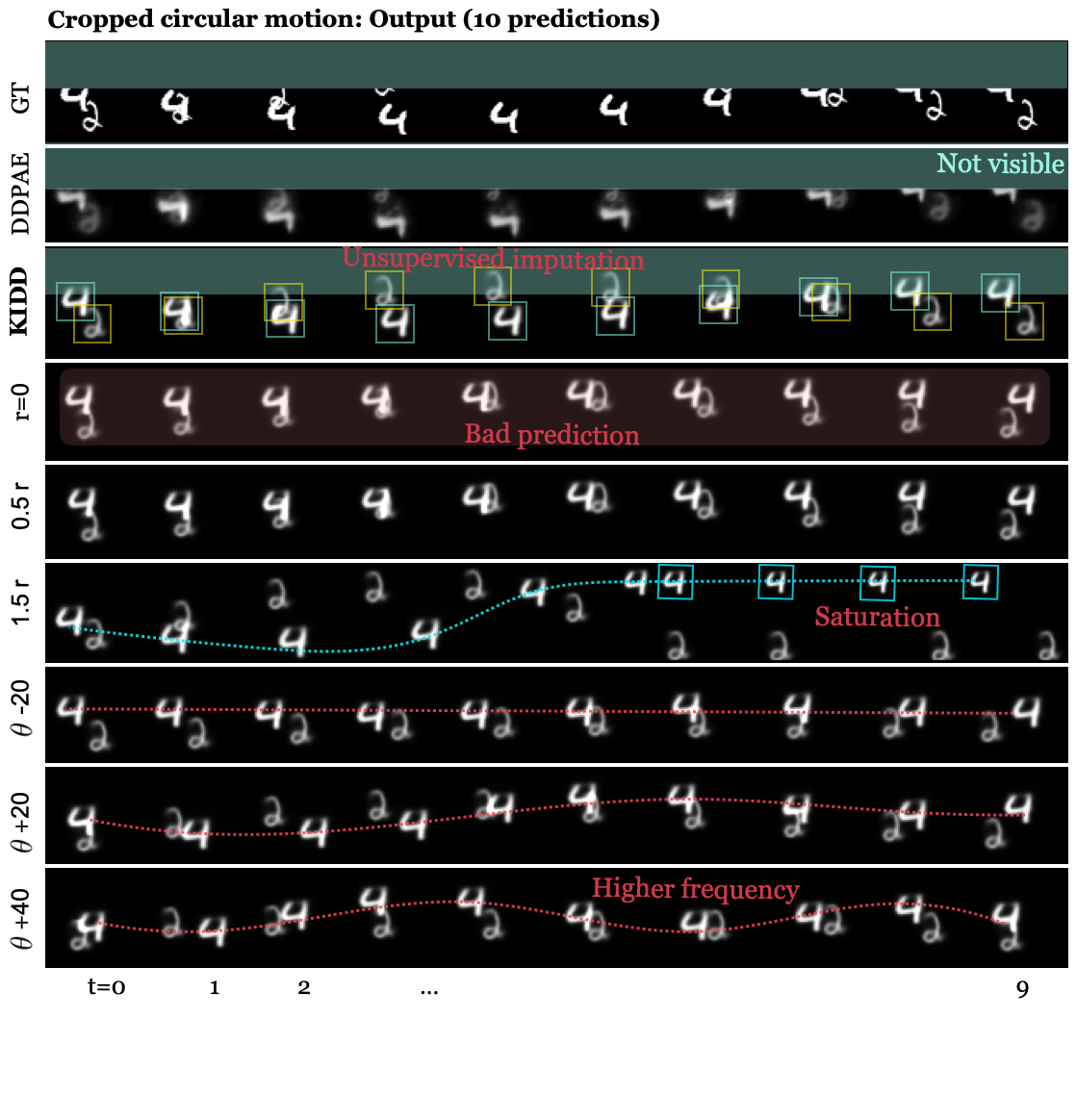}\\
\vspace{-5mm}
\caption{Ablation study and qualitative results of the \textit{Cropped circular motion} scenario. In transparent blue, we show the occluded area of the frame. The labels in the Y axis indicate the variations sufferd by a selected eigenvalue of the learned $\mathcal{K}$. Written in red, we indicate the behaviors we perceive. In this particular case it's interesting to highlight the unsupervised imputation of the trajectory that \ours{} discovers in an unsupervised fashion. This indicates that the model learns the correct dynamics. \textsc{ddpae} fails to reconstruct the objects properly.}\label{fig:crop-app}
\end{figure*}

\begin{figure*}[t]
\centering
\includegraphics[height=0.36\textwidth]{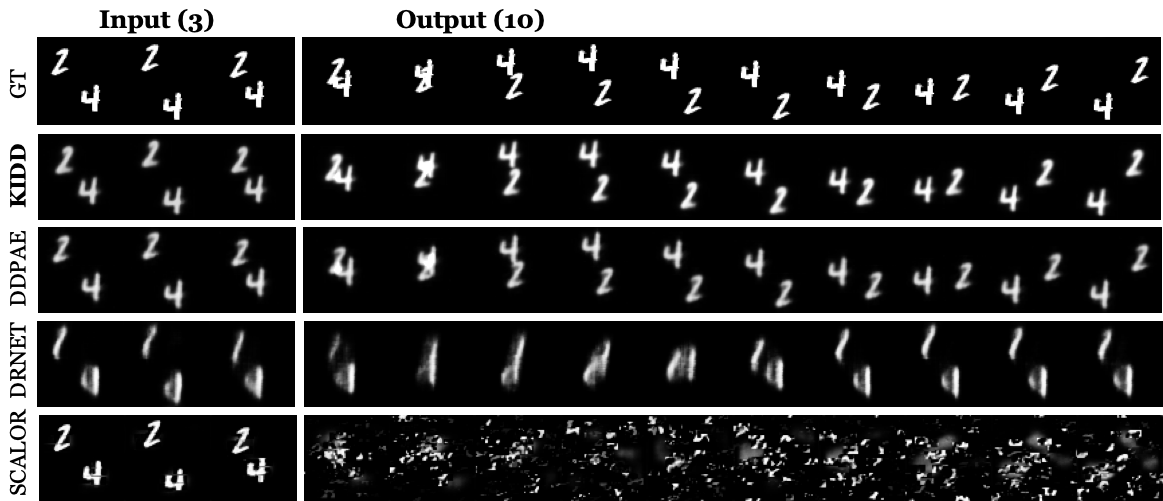}\\
\includegraphics[height=0.36\textwidth]{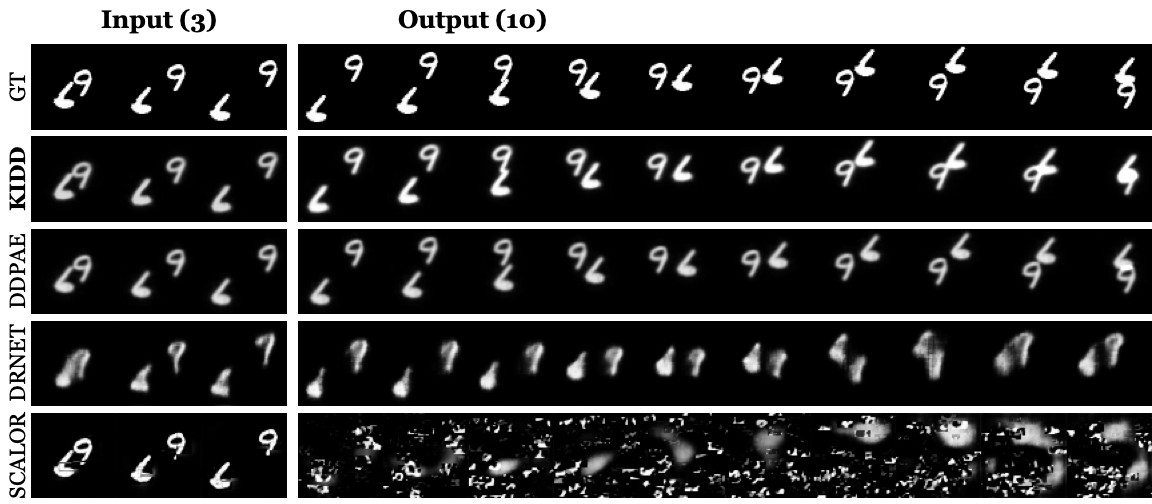}\\
\includegraphics[height=0.36\textwidth]{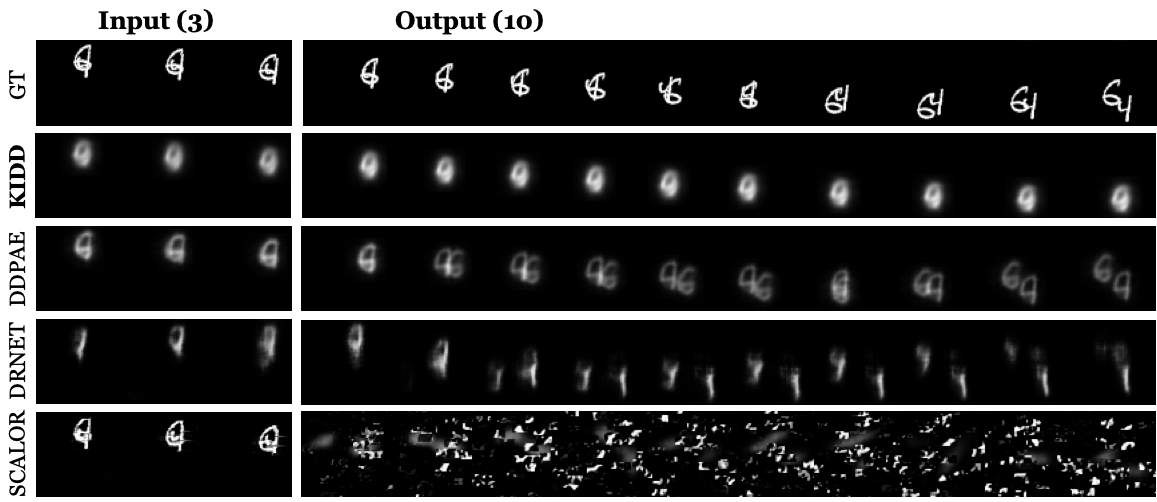}
\caption{Qualitative comparison (with failure cases) of \ours{} against the baselines for \textit{Inelastic/Superelastic collision} case. In the top, success case where the inelastic collision is properly modelled by \ours{}. The best baseline, \textsc{ddpae}, also reconstructs the trajectory successfully. In the center, failure case where the decomposition and reconstruction of the scene is correct, but the dynamics are slightly off for \ours{}, from the points of collision. In the bottom, failure case in which \ours{} is unable to decompose the scene. Note that \textsc{scalor} provides a very poor prediction although it has the best reconstruction.}\label{fig:cte-app}
\end{figure*}

\begin{figure*}[h]
\centering
\includegraphics[height=0.17\textwidth]{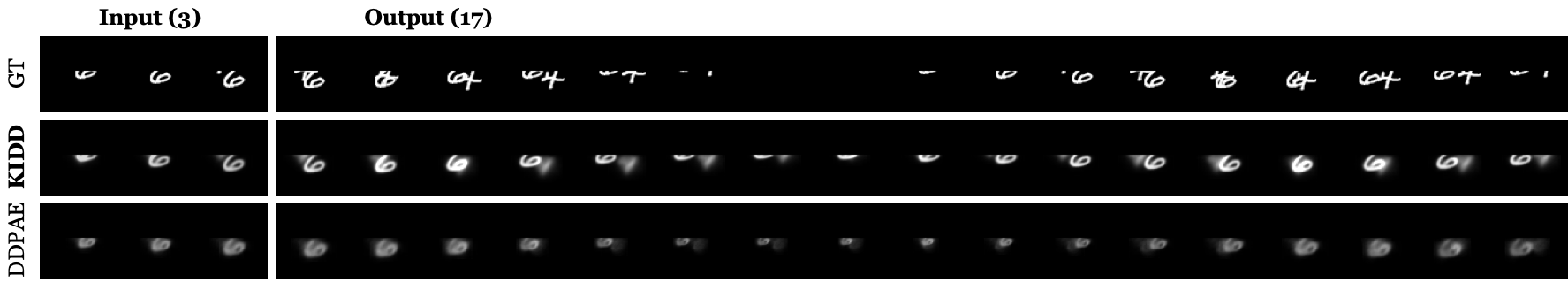}\\
\caption{Failure case for the \textit{cropped circular motion} scenario. In this case, one of the numbers is never seen in the initial conditions, and therefore \ours{} can't reconstruct its appearance. Note that \textsc{ddpae} is also unable to reproduce it, and the results are qualitatively worse.}\label{fig:crop-app-bad}
\end{figure*}
\section{More examples and failure cases} \label{app:extra-exp}

In this section, we provide examples for the four studied scenarios, including failure cases for \ours{}. We will also show more examples of our ablation studies. For the latter, we modify the eigenvalues of the learned Koopman operator $\mathcal{K}$ to study and visualize the variations in the dynamics of the objects in the scene. Figure \ref{fig:eig-app} gives an overview of the learned eigenvalues for three of the studied scenarios: \textit{Circular motion, Cropped circular motion} and \textit{3D to 2D motion projection}. For these cases, there are no inputs $\V{u}_t^i$, and therefore the operator $\mathcal{K}$ generates the dynamics. 

Figure \ref{fig:eig-app} shows shadowed in red the eigenvalues that have been empirically proven to have negligible effect on the performance in Table 1. For all cases, 6 eigenvalues (usually 3 complex conjugate pairs) are enough to generate the behaviour seen in the dynamics of the Koopman manifold. We chose an eigenvalue pair (highlighted in green) and changed its radius $r$ and angle $\theta$ to study the effects on the scene dynamics.

In Figure \ref{fig:sph-app}, we see the case of \textit{3D to 2D motion projection}. Here, we display the two best-performing RNN-based baselines (\textsc{ddpae} and \textsc{drnet}) toghether along \ours{}. Qualitatively, we see that they perform similarly or worse than \ours{} in this case. We can see how the model identifies and predicts independently each one of the digits. The blue dotted line shows behaviors due to changes in magnitude of the eigenvalue pair. When the eigenvalues are outside of the unit circle, the system they model is unstable. This can be observed in the figure by looking at the behaviour for $\hat{r}=1.5r$. The digits start showing an unstable behaviour by changing progressively it size and the amplitude of their oscillation. At a certain point, the digit gets stuck in the bottom frame limit, reaching the constraint of the pose vector.
For variations in the phase angle $\theta$, it is interesting to note that it has a direct link to the frequency of the digit's oscillation. When $\hat{\theta} = \theta - 20$, $\hat{\theta}$ is close to 0. This has a clear impact on the vertical component of the object's trajectory. When we increase $\theta$, the vertical oscillation frequency increases with it. These behaviors are as expected given the illustration in Figure 3
(top-left).

We see a very similar behavior in Figures \ref{fig:circ-app} and \ref{fig:crop-app}. For $\hat{r}>1$, Figure \ref{fig:circ-app} shows an oscillating behavior of the digit sizes that increase with $t$. Again, the behavior seems unstable. Figure \ref{fig:crop-app} illustrates a familiar saturation behavior for $\hat{r}>1$. Also, it shows how \ours{} is able to find the unseen dynamics of a digit in a partially visible frame. If we observe the third row of Figure \ref{fig:crop-app}, we can see how the digit ``2" has the expected oscillation, even when it was unseen neither in the input data or the self-supervision. This is an indicator of the ability of \ours{} to discover the true dynamics of a system. It also exhibits better reconstruction and prediction that the strongest baseline \textsc{ddpae}.

Figure \ref{fig:cte-app} shows three particular cases of the \textit{Inelastic/Superelastic collision}. The first  example is a success, the second one is a partial success and the third one is a failure. In all cases we see a similar behaviour for the baselines. \textsc{ddpae} is the closest to our model in capturing the dynamics for prediction, with the difference that it leverages RNNs. \textsc{scalor} has a very good reconstruction, but its prediction is not comparable to the other tested architectures. \textsc{drnet} seems to decompose the scene and capture partially the dynamics and appearance. However, its performance is poor.
For the top case, \ours{} correctly decomposes the scene and predicts the inelastic and superelastic collisions with the result of an accurate and sharp prediction. For the center case, the performance is similar. However, we can observe that the final digits are a bit off from their actual trajectory. This is likely due to a bad modelling of the collisions. Finally, in the bottom example all methods fail to decompose the scene into its composing objects, as they appear overlapped. 
\\
\\
Finally, Figure \ref{fig:crop-app-bad} illustrates a failure case for the \textit{Cropped circular motion} scenario. In this case, the input frames are heavily occluded. One of the digits (6) is fully visible but the other is not. In this case, as expected, none of the methods can model the occluded digit. However, given the uncertainty, \ours{} has a sharper and more accurate prediction than \textsc{dddpae}.

\end{document}